\documentclass[runningheads]{llncs}
\usepackage{graphicx}
\usepackage{subfigure}
\usepackage{comment}
\usepackage{amsmath,amssymb} 
\usepackage{color}
\usepackage{array}
\usepackage{multirow}
\usepackage{booktabs}
\usepackage{subfigure}
\usepackage{algorithm}
\usepackage{algorithmic}
\def\ie{\emph{i.e.}}
\def\eg{\emph{e.g.}}
\def\etc{\emph{etc.}}
\newcommand{\etal}{\textit{et al.}}
\usepackage[breaklinks=true,letterpaper=true,colorlinks,bookmarks=false]{hyperref}

\begin{document}
\pagestyle{headings}
\mainmatter

\title{SSN: Shape Signature Networks for Multi-class Object Detection from Point Clouds} 


\titlerunning{SSN: Shape Signature Networks}
%
\author{Xinge Zhu$^{\dag}$~~~~~ Yuexin Ma$^{\S}$~~~~~ Tai Wang$^{\dag}$~~~~~ Yan Xu$^{\dag}$\\ Jianping Shi$^{\ddagger}$~~~~~ Dahua Lin$^{\dag}$ }
\institute{$^{\dag}$The Chinese University of Hong Kong~~~~$^{\ddagger}$SenseTime Research \\ $^{\S}$Hong Kong Baptist University\\ {\tt\small {\{zx018, wt019, xy019, dhlin\}@ie.cuhk.edu.hk}} \\ {\tt\small yuexinma@comp.hkbu.edu.hk, shijianping@sensetime.com}}
\authorrunning{X. Zhu, Y. Ma, T. Wang, Y. Xu, J. Shi, D. Lin}
%
\maketitle

\begin{abstract}

Multi-class 3D object detection aims to localize and classify objects of multiple categories from point clouds. 
Due to the nature of point clouds, \ie~unstructured, sparse and noisy, some features benefitting multi-class discrimination are underexploited, such as shape information.
In this paper, we propose a novel 3D shape signature to explore the shape information from point clouds. By incorporating operations of symmetry, convex hull and chebyshev fitting, the proposed shape signature is not only compact and effective but also robust to the noise, which serves as a soft constraint to improve the feature capability of multi-class discrimination.
Based on the proposed shape signature, we develop the shape signature networks (SSN) for 3D object detection, which consist of pyramid feature encoding part, shape-aware grouping heads and explicit shape encoding objective.
Experiments show that the proposed method performs remarkably better than existing methods on two large-scale datasets. Furthermore, our shape signature can act as a plug-and-play component and ablation study shows its effectiveness and good scalability.\footnote{Source code: \url{https://github.com/xinge008/SSN}}

\end{abstract}

\section{Introduction}

The success of autonomous vehicles in urban scene heavily relies on the ability to handle the complex environments, where the accurate and robust perception is the foundation. 
To achieve this, autonomous vehicles are equipped with various sensors, including camera, radar and lidar, in which lidar is considered as the most critical one.
The lidar sensor could provide the accurate depth information which is a significant advantage than image and thus lidar-based object detection~\cite{second,pointpillar,voxelnet,hdnet} also achieves greatly better performance than image-based methods~\cite{chen2016monocular,pseudo,stereo2,liu2019deep}. The mainstream 3D detection frameworks often focus on the single-category detection, such as car or pedestrian, while in the real world the autonomous vehicles need to detect multi-class objects simultaneously. In this way, how to distinguish heterogeneous categories plays an indispensable role in the success of multi-class 3D object detection.

\begin{figure}[t]
\begin{center}
   \includegraphics[width=0.9\linewidth]{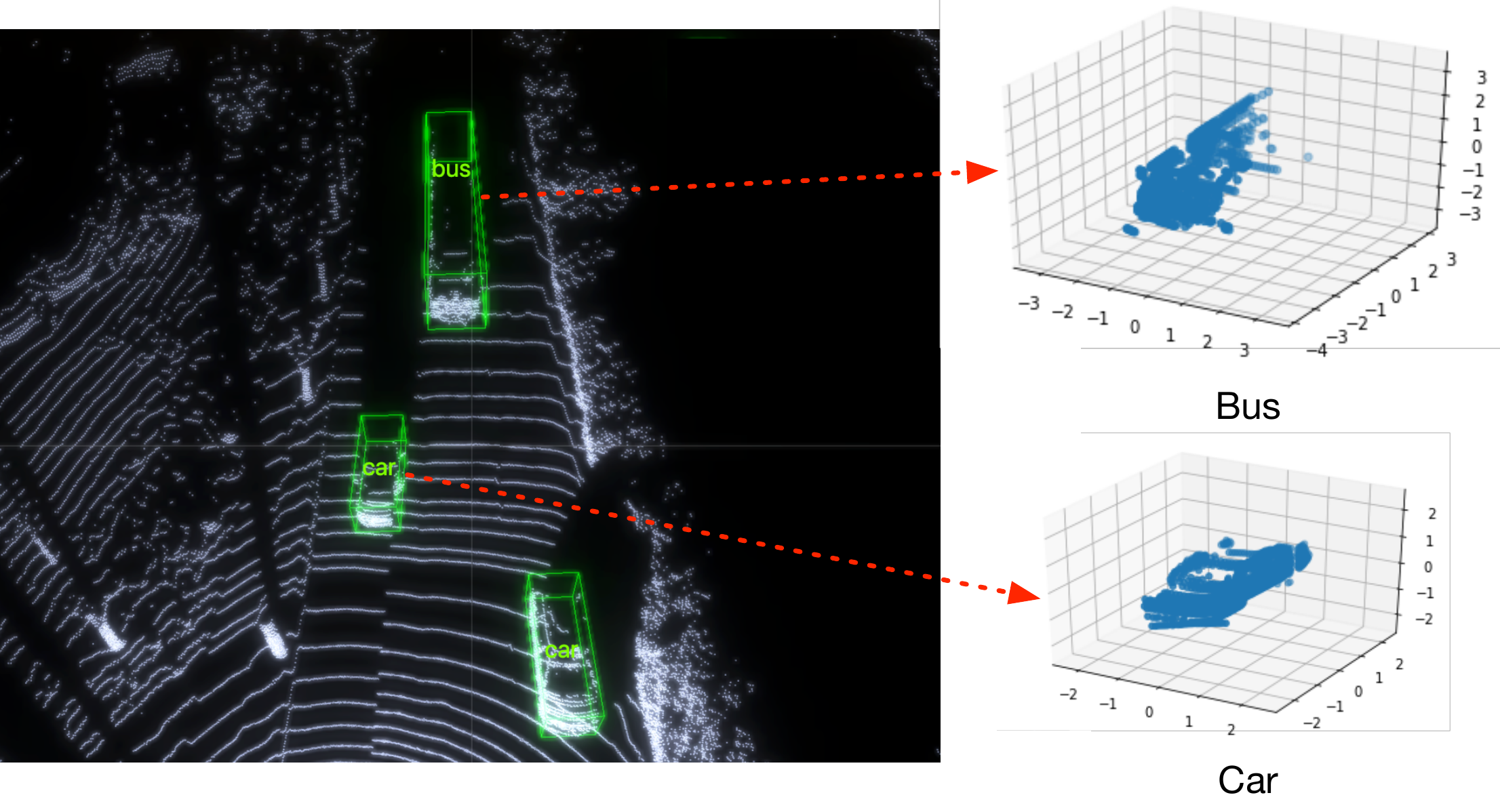}
\end{center}
\vspace{-3ex}
   \caption{\small
   We show an example of the point clouds of two objects belonging to different categories. It is noted that they have different shapes and scales.} 
   \label{fig:teaser}
\end{figure}

A natural idea to handle this challenge is to utilize the difference on appearance or texture to distinguish different objects. Unfortunately, this approach is not feasible for point clouds, due to its point-based representation lacking of texture or appearance. An appealing alternative is to explore the shape information to guide the discriminative feature learning. Fig.~\ref{fig:teaser} shows an example that demonstrates the shape difference between two categories. From the teaser, we can find that the shape and scale vary with the categories. However, due to the sparsity and noise of the point cloud, how to build the effective and robust shape encoding remains a widely open question. 

In this paper, we propose a novel shape signature for shape encoding, which possesses two appealing properties, \ie~{\textbf{compact}} (effective and short as the objective) and {\textbf{robust}} (robust against the sparsity and noise).
Specifically, as the scan of lidar often covers parts of object (\eg~two or three faces), we first use the {\emph{symmetry}} operation to complete the sparse points. Then we {\emph{project}} the points to three views of the object, including bird view, side view and front view, for thoroughly modeling the shape information. Furthermore, the {\emph{convex hull}} is introduced to represent the shape of three views, making it robust to the inner-sparsity. Based on the convex hull, we use an {\emph{angle-radius strategy}} to form the function of convex hull, in which each separate angle corresponds to a radius from inner-center to contour. Finally, to make the shape encoding compacter and more robust, we apply the {\emph{Chebyshev fitting}} to perform the approximation on the function of angle-radius strategy, and then the final shape encoding is formed by the coefficients of Chebyshev approximation. Note that the proposed shape signature aims to keep the shape information \textbf{consistent} (not same) within the same category and separate the shape distributions across different categories, which enables the shape signature serve as a soft constraint for learning discriminative and class-specific features. 

Based on the proposed shape signature, we develop the shape signature networks for multi-class 3D object detection. The basic idea is to incorporate the shape information to better distinguish multiple categories. Specifically, SSN consists of four components, point-to-structure, pyramid feature encoding part, shape-aware grouping heads and shape signature objective. Here, shape-aware grouping heads bring the objects with similar shape together, so as to share weights based on the object size (\eg~bus and truck need a heavier head than car); while shape signature acts as an auxiliary objective, thus benefitting the feature capability of multi-category discrimination. 

We tested the proposed framework on two large-scale datasets, nuScenes~\cite{nuscenes} and Lyft~\cite{lyft} dataset, which contain multi-class objects, including car, bus, pedestrian, motorcycle and \etc~On these experiments, SSN yields considerable improvement over existing methods, about {\bf{10}}\% in NDS and {\bf{5}}\% in mAP-3D, respectively. 
We also make an in-depth investigation on the proposed shape signature, showing its good scalability with different backbone networks on different datasets. TSNE visualization of shape signature vector also verifies its role of soft constraint.

The contributions of this work mainly lie in four aspects:
(1) We propose a novel shape signature to explicitly explore the 3D shape information from point clouds, which is compact but contains sufficient information, and robust against the noise and sparsity.  
(2) We develop the shape signature networks (SSN) for object detection from point clouds,
which effectively perform the multi-class detection through shape-aware heads grouping and shape signature investigation.
(3) We conduct extensive experiments to compare the proposed methods with
others on various benchmarks, where it consistently yields notable performance gains.
(4) The proposed 3D shape signature could act as a plug-and-play component and be independent to the backbone. Experiments on different backbone networks show its good scalability.

\section{Related Work}
\noindent \textbf{Shape Representation}~~~
Numerous works processing on this research area have been made in recent decades.
Johnson \etal ~\cite{spin} introduced a local shape based descriptors on 3D point clouds called spin images. Based on spin image, Golovinskiy \etal ~\cite{spin2} incorporated the contextual features into shape descriptor. While these local descriptors construct encoding resorting to the local neighborhood, global descriptors~\cite{belongie2001shape,frome2004recognizing,good,marton2011combined} encode the geometric and structured information of the whole 3D point cloud. 
IS~\cite{IS} introduced an implict shape signature for instance segmentation by using the auto-encoder to learn a low-dimensional shape embedding space.
Viewpoint Feature Histogram (VFH)~\cite{rusu2010fast} used the viewpoint direction component and surface shape component to bin the point cloud for shape encoding. However, most of them do not pursue the compact representation and the robustness to the sparsity, which is the major difference between our shape signature and theirs. The proposed shape signature performs the symmetry for completion, convex hull for inner sparsity and Chebyshev fitting for short vector. The cooperation of these operations leads to the compact and robust shape encoding.

\noindent \textbf{3D Object Detection}~~~
Most 3D object detection methods can be divided into two groups: image-based methods and lidar-based methods. For the image-based methods, the key insight is to estimate the reliable depth information to replace lidar~\cite{pseudo,stereo2,liu2019deep}. Monocular or stereo based depth estimation methods~\cite{xu2019depth} have greatly pushed forward the-state-of-art in this field. \cite{mono2} introduced a multi-level fusion method by concatenating the image and generated depth map. \cite{mono5} incorporated depth features including disparity map and distance to the ground into the detection framework.
However, although the image-based methods have made significant progress, the performance of this type of methods still lags far behind lidar-based methods.

Lidar-based methods are the mainstream of 3D detection task as lidar provides accurate 3D information. Most lidar-based methods process the unstructured point input in different representations. In~\cite{voxelnet,second,wangrecon}, point cloud were converted into voxels and a SSD~\cite{ssd} based convolution network was used for detection. PointPillar~\cite{pointpillar} used the pillar to encode the point cloud with PointNet~\cite{pointnet}. \cite{mv3d,pixor,defu,ku2018joint,Xu2017PointFusionDS,Liang2019MultiTaskMF} converted point cloud data into a BEV representation and then fed them into the structured convolution network. \cite{pointrcnn,std,qi2019deep} introduced the two-stage detector into 3D detection, where coarse proposals were first generated and then refine stage was used to get the final predictions. \cite{frustum} used the raw point cloud as input and extracted the frustum region reasoned from 2D object detection to localize 3D objects. However, most of them focus on the single-class detection, while neglecting to explore the multi-class discrimination. Compared to these works, our proposed method differs essentially in that it effectively explores the shape information, which plays a crucial role in distinguishing multi-class objects.

\section{Methodology}

\subsection{Overview}
Given a point cloud, our goal is to localize and classify the multi-class target objects. Unlike the single-class detectors, we desire to obtain a detector which could effectively distinguish the objects from multiple categories. To this end, we propose a multi-class 3D detection framework based on shape information exploration. The basic idea is to utilize the shape information via two key ingredients, \ie~shape signature objective and shape-aware grouping heads, to benefit the multi-class classification. 

As shown in Fig.~\ref{fig:pileline}, our framework consists of four components, \ie~point-to-structure, pyramid feature encoding, shape-aware grouping heads and multi-task objectives, where point-to-structure and pyramid feature encoding are flexible (\ie~multiple options are available). The key components of SSN are the shape signature objective and the shape-aware grouping heads. 
Particularly, during the training the shape signature objective could guide the learning of discriminative features via back-propagation, benefitting the multi-class discrimination. After training, the shape signature objective is no longer needed. In what follows, we will present the details of shape signature and SSN.

\subsection{Shape Signature}
Given the ground truth points of object, we parameterize the shape information of the object with the proposed shape signature, then apply the obtained shape signature vector as a soft constraint to improve the feature capability of multi-class discrimination.
As mentioned above, the desired shape signature should carry two properties: 1) compact and effective as a part of objective; 2) robust to the sparsity and noise. To achieve this, we introduce several operations to handle the issue of point clouds. As shown in Fig.~\ref{fig:ss}, the shape signature contains two components, shape completion and shape embedding, where shape completion consists of Transform and Symmetry, and shape encoding involves Projection, Convex Hull, Angle-Radius and Chebyshev Fitting. 

\vspace{-1ex}
\subsubsection{Shape Completion}
Since the scan of lidar sensor only covers the partial observation, this property limits the shape investigation. We thus introduce the shape completion to tackle this issue, which consists of following steps.

\noindent \textbf{Transform.}~~~ The points of target object are located in the scene. We first transform the center of ground truth box to the origin point, and use the forwarding direction as the reference axis.

\begin{figure*}[t]
\begin{center}
   \includegraphics[width=1.0\linewidth]{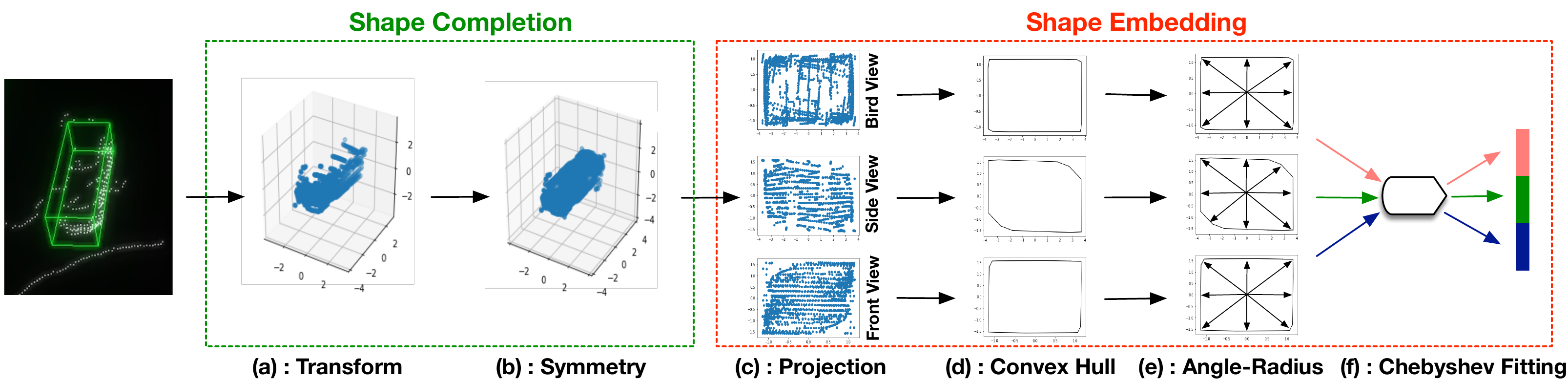}
\end{center}
\vspace{-3ex}
   \caption{We show the workflow of the proposed shape signature. Two major components, \ie~Shape Completion and Shape Embedding, are illustrated with two dashed rectangles. Specifically, step (a) is to transform the center of box to the origin point. (b) is the symmetry for completing the partial observation. (c) is to project the 3D points into three views. (d) is to extract the convex hull to enhance the robustness to sparsity. (e) is the Angle-Radius and step (f) is the Chebyshev fitting to get the final shape vector.
   (\textbf{Best viewed in color)}.}
\label{fig:ss}
\vspace{-3ex}
\end{figure*}

\noindent \textbf{Symmetry.}~~~ Lidar scans could only cover two or three faces of object, thus this partial observation would affect the investigation of shape. We introduce the centro-symmetry to complete the partial view. From Fig.~\ref{fig:ss} (b), we can find that after symmetry, the points of target object become more dense and the observation gets complete. 

\vspace{-1ex}
\subsubsection{Shape Embedding} We then introduce following operations to achieve the compact and effective shape embedding.

\noindent \textbf{Projection.}~~~ Given the completed points, we project the 3D points to three 2D views, \ie~bird view, front view and side view. 
Based on the projection, the 3D points are decoupled into several 2D planes, which could thoroughly describe the 3D shape and benefit the reduction of parameters.

\noindent \textbf{Convex Hull.}~~~ After projection, we get 2D points of different views. However, it can be found that the organization of 2D points is limited to effectively represent the shape and there still exists the inner-sparsity. Hence, the convex hull is introduced to characterize these 2D points and emphasize the contour of views, thus being robust to the inner-sparsity. Furthermore, the contour of 2D points also maintains the scale information, which is an important factor for multi-class discrimination (see Fig.~\ref{fig:ss} (d)).

\noindent \textbf{Angle-Radius.}~~~ To describe the convex hull and highlight the contour shape and scale, we design an angle-radius parametric function $f(\theta)$. We use the center of ground truth box as the origin point $\sigma$ and densely sample some angles $\theta$. In this way, the function 
$f(\theta) = dist(\sigma \stackrel{\theta}{\longrightarrow} \mathbb{C})$, where $\mathbb{C}$ is the convex hull and $dist$ indicates the distance between origin point and intersection point (\ie~radius). In the implementation, we sample 360 angles and calculate the radius accordingly. 

From Fig.~\ref{fig:ss} (e) (see the aspect ratios), it is noted that the function $f(\theta)$ involving the angle and radius does well in maintaining the shape and scale of contour. However, the dense sampling also introduces the long vector (360 dimensions) which is not desired for the objective. Hence, to shorten the long vector representation and further enhance the robustness against the noise (\eg~some outliers in the 2D points), we introduce the Chebyshev Fitting to process the angle-radius function $f(\theta)$.

\noindent \textbf{Chebyshev Fitting.}~~~ Chebyshev Polynomials Fitting~\cite{cheb} provides an approximation that is close to the polynomial of best approximation to a function under the maximum norm. Our goal is to apply the Chebyshev polynomials to approximate the angle-radius function, and then use their coefficients to serve as the final shape vector. 

There are two kinds of Chebyshev polynomials fitting~\cite{cheb}, and we use the Chebyshev polynomials of first kind. The first kind $T_n(x)$ is defined by the recurrence relation:
\begin{align}
T_0(x) &= 1, T_1(x) = x,\\
T_{n+1}(x) &= 2xT_n(x) - T_{n-1}(x).
\end{align}
Hence, the generic formulation of Chebyshev approximation can be written as a sum of {$T_n(x)$}.
\begin{align}
f(x) \approx \sum_{n=0}^{N} \alpha_n T_n(x),
\end{align}
where $\alpha$ are the coefficients. These coefficients can be computed with the formulas:
\begin{align}
\alpha_0 &= \frac{1}{N+1}\sum_{n=0}^{N}f(x_n)T_0(x_n)\\
\alpha_j &= \frac{2}{N+1}\sum_{n=0}^N f(x_n)T_j(x_n)
\end{align}
Since the number of coefficients in $f(x)$ is $2^{N-1}$, we truncate $\alpha$ with top $k$ terms. For each view, top $k$ coefficients are the shape vector. The final shape signature is $ [\underbrace{\alpha_1, \dots, \alpha_k}_{\text{Birdview}}, \underbrace{\alpha_1, \dots, \alpha_k}_{\text{Sideview}}, \underbrace{\alpha_1, \dots, \alpha_k}_{\text{Frontview}}]$. In the implementation, we use $k$=3 and the dimension of final shape signature vector is 9, which is suitable to serve as an objective for the network.

\noindent\textbf{Some Extreme Cases.}~~~ Due to the limitation of Lidar sensor and human annotators, some ground truth boxes contain less than or equal to 5 points, even 0 point for incorrect labeling. For these boxes, it is hard to model the shape information, and we thus use the average encoding of that category to represent their shape vectors.

\begin{figure*}[t]
\begin{center}
   \includegraphics[width=1.0\linewidth]{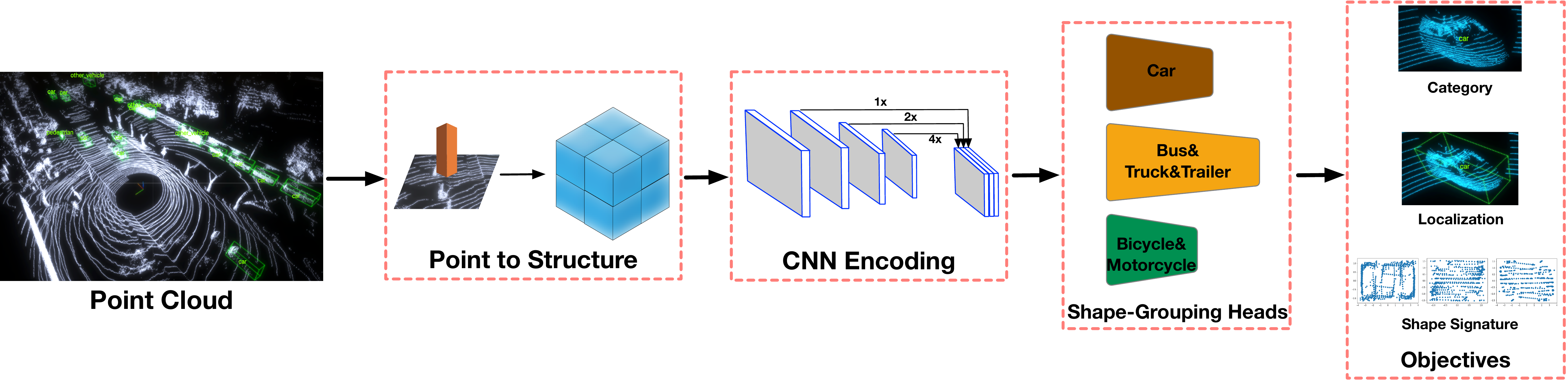}
\end{center}
\vspace{-2ex}
   \caption{The pipeline of our framework SSN. Four major components are illustrated with four dashed rectangles. The first one is the Point-to-Structure part, which converts the raw points into the structured representation, such as voxels~\cite{second,voxelnet} or pillars~\cite{pointpillar}. The second is the pyramid feature encoding part. The third one is the shape-aware grouping heads, which consist of multiple branches for objects with similar shape and scale. The final part is the objective, including classification, localization and shape signature regression.
   (\textbf{Best viewed in color}).}
\label{fig:pileline}
\end{figure*}

\subsection{SSN: Shape Signature Networks}
Based on the proposed shape signature, we design the SSN to achieve the effective multi-class 3D detection. We first describe each component, especially two key ingredients, \ie~shape-aware grouping heads and shape signature objective, then we integrate different parts to form the unified target: exploring the shape information to better distinguish multi-class objects.

\vspace{1ex}
\noindent\textbf{Point-to-Structure.}~~~ Since the organization of point cloud is unstructured, the first step is to transform the point cloud to the structured representation. As mentioned above, multiple options are available in this part, such as the voxel-based~\cite{second,voxelnet} representation or pillar-based~\cite{pointpillar} or Bird-view representation~\cite{mv3d}. After obtaining the structured representation, the subsequent 2D convolution or 3D convolution networks can be applied. In the implementation, we choose the pillar-based representation to structure the point clouds. Furthermore, we also test the shape signature with other structure representation (voxel-based) and the proposed shape signature shows good scalability.

\vspace{1ex}
\noindent\textbf{Pyramid Feature Encoding.}~~~ We follow the idea of FPN~\cite{FPN} to perform the feature encoding. A top-down convolutional network is first applied to extract the feature from multiple spatial resolutions. Then all features are fused together through upsampling and concatenation.

\noindent\textbf{Shape-aware Grouping Heads.}~~~
Since multi-class target objects vary significantly in scale and shape, we propose the shape-aware grouping heads to adapt this ideology for multi-class discrimination. The basic idea is to create multiple heads, in which objects with similar scale and shape share the weights. The reasons mainly lie in the following: 1) objects with different scale and shape should have different heads. For example, the head of bus needs to be heavier (or more deep) than the head of bike due to its large scale, because heavier head, larger receptive field. 2) shape grouping heads could perform the coarse shape exploration and also alleviate the effect from other groups.

As shown in Fig.~\ref{fig:pileline}, the design of shape-aware grouping heads follows the spirit of ``larger object, heavier head". Based on the shape and scale of target objects, we group the bus, truck and trailer together with a heavier head, and gather bicycle and motorcycle with a lighter head, and treat the car with a medium head. Each head only covers the prediction of corresponding categries. By integrating above components, a SSD-based detection framework is formed.

\subsection{Multi-task Objectives}
In our framework, there are three objectives, \ie~multi-class classification, localization regression and shape vector regression. For the multi-class classification, we follow the previous work~\cite{second} to use the focal loss~\cite{focal}
\begin{align}
\mathcal{L}_{cls} = -\alpha_t(1-p_t)^{\gamma}\log(p_t),
\end{align}
where $p_t$ is the class probability of the default box and we use $\alpha=0.25$ and $\gamma=2$. 

For the localization loss, we use the smooth L1 loss to minimize the distance between predictions and localization residuals~\cite{second}. 
\begin{align}
\mathcal{L}_{loc} = \text{SmoothL1}(\triangle b),
\end{align}
where $\triangle b$ are the localization residuals, including the center ($x,y,z$), scale ($w,h,l$) and rotation ($\theta$).

Unlike regressing the residuals in localization, the network is trained to directly regress the shape vector. For the shape regression, we also apply the smooth L1 loss.
\begin{align}
\mathcal{L}_{shape} = \text{SmoothL1}(\mathbb{S}),
\end{align}
where $\mathbb{S}$ is the shape vector.

The total objective of three tasks is therefore:
\begin{align}
\mathcal{L} = \beta_1 \mathcal{L}_{cls}  + \beta_2 \mathcal{L}_{loc} + \beta_3 \mathcal{L}_{shape},
\end{align}
where $\beta$ are the constant factors of loss terms. As the shape loss is much larger than localization and classification loss, we set $\beta_1 = 1.0$, $\beta_2 = 1.0$  and $\beta_3 = 0.5$ to balance the value scale.

\section{Experiments}
\subsection{Datasets}

Two large-scale datasets, nuScenes dataset and Lyft dataset, are applied in experiments. The details of two datasets are shown in the following.

\noindent{\textbf{NuScenes Dataset}}~\cite{nuscenes}~~~ It collects 1000 scenes of 20s duration with 32 beams lidar sensor. The number of total frames is 40,000, which is sampled at 2Hz, and total 3D boxes are about 1.4 million. 10 categories are annotated for 3D detection, including Car, Pedestrian, Bus, Barrier, and \etc (details in the experimental results). They also officially split the data into training and validation set, and the test results are evaluated at EvalAI\footnote{\url{https://evalai.cloudcv.org/web/challenges/challenge-page/356/overview}}. Furthermore, a new metric is also introduced in nuScenes dataset, namely nuScenes detection score (NDS)~\cite{nuscenes}, which quantifies the quality of detections in terms of average classification precision, box location, size, orientation, attributes, and velocity. The mean average precision (mAP) is based on the distance threshold (\ie~ 0.5m, 1.0m, 2.0m and 4.0m).
The whole range is about 100 meter, and we mainly use the range of 0-50m in full 360 degree. More detailed descriptions are shown in Supplementary Materials.

\noindent{\textbf{Lyft Dataset}}~\cite{lyft}~~~ It contains one 40-beam roof lidar and two 40-beam bumper lidars, and in the experiments, we only use the data from roof lidar. The data format is similar to the nuScenes dataset. Total 9 categories are annotated for detection, including car, emergency\_vehicle, motorcycle, bus, truck, and \etc.
Total 22,680 frames are used as the training data, and test set contains 27,468 frames while 30\% of the test data is for validation in Kaggle competition\footnote{\url{https://www.kaggle.com/c/3d-object-detection-for-autonomous-vehicles}}. The evaluation metric is the mean average precision, which is similar to the metric of COCO dataset but calculates the 3D IoU (with the threshold of 0.5, 0.55, 0.6, 0.65, ..., 0.95). Hence, we name it as \textbf{mAP-3D}, and it is worthy to note that mAP-3D is much strict than mAP in nuScenes and Kitti~\cite{kitti}.

\subsection{Implementation Details}
In our implementation, we use the pillar based~\cite{pointpillar} method to convert the point cloud to the structured representation. 
For nuScenes dataset, the x, y, z range is ([-49.6, 49.6], [-49.6, 49.6], [-5, 3]) and the pillar size is [0.2, 0.2, 8]. The max number of pillars is 30,000 and max number of points per pillar is 20.
For Lyft dataset, the range is ([-89.6, 89.6], [-89.6, 89.6], [-5, 3]) and the pillar size is [0.2, 0.2, 8] too. The max number of pillars is 60,000 and max number of points per pillar is 12. 

For the anchors, we calculate the mean width, length and height of each class and use birdview 2D IoU (width and length) as the matching metric; when the matching between anchors and ground truth is larger than the positive threshold, these anchors are positive, otherwise if the matching is smaller than negative threshold, they are negative anchors. The matching threshold is different for different categories.
During inference, the multi-class and rotational NMS is employed, where multi-class NMS indicates applying NMS for each class independently. For a \textbf{fair comparison}, no multi-scale training / testing, SyncBN and ensemble are applied. For nuScenes dataset, online ground truth sampling~\cite{second,Zhu2019ClassbalancedGA} is not used. 
We also submit the results on these official websites~\footnote{\url{https://www.nuscenes.org/object-detection}}\footnote{\url{https://www.kaggle.com/c/3d-object-detection-for-autonomous-vehicles}} (Our submissions are ``gezi" and ``OIDH" respectively, and both of them are anonymous.).

\noindent{\textbf{Network Details}}~~~
For the point-to-structure, we follow the network in~\cite{pointpillar}, where a simplified PointNet is used. It contains a linear layer, BatchNorm and ReLU layer to handle the features of pillars.
For the CNN feature encoding, the FPN based module is introduced to extract the fused features. Three levels of features are first upsampled with the transposed 2D convolution, and then concatenated. 
For the shape-aware grouping heads, objects with similar shape and scale share the same head. For bus, truck and trailer, a heavier head is applied, where two downsample blocks process the features from FPN. Each downsample block consists of 3x3 2D convolution layer with stride=2, followed by BatchNorm and ReLU. For the lighter head (such as bicycle, motorcycle), the block with stride=1 is used. For the medium head, one downsample block is applied. Note that another block with stride=1 is followed in each downsample block. More detailed network structure is shown in Supplementary Materials. 

\noindent{\textbf{Optimization}}~~~ We use the Adam optimizer with cycle learning decay. The maximum learning rate is 3e-3 and weight decay is 0.001. We train 60 epoches and 80 epoches for nuScenes dataset and Lyft dataset, respectively; the batch size is 2 for nuScenes and 1 for Lyft dataset.

\subsection{Results}

\begin{table*}
\small
\caption{Results of multi-class 3D detection on nuScenes dataset. ``Trail", ``CV", ''Ped" , ``MC", ``Bicy", ``TC", ``Bar" indicates the trailer, construction vehicle, pedestrian, motorcycle, bicycle, traffic cone, and barrier respectively. Bold-face and underline numbers denote the best and second-best respectively }
\vspace{-1ex}
\setlength{\tabcolsep}{0.4pt}
\begin{center}
\begin{tabular*}{1.0\linewidth}{c|c|c| c| c| c| c| c| c| c|c|c|c|c}
\toprule
Methods & Modality & Car & Truck & Bus & Trail & CV & Ped & MC & Bicy & TC & Bar & mAP  &NDS\\
\hline
\hline
Mono~\cite{dism} & RGB & 47.8 & 22.0 & 18.8 & 17.6 & 7.4 & 37.0 & 29.0 & \textbf{24.5} & 48.7 & 51.1 & 30.4  & 38.4\\
\hline
Second~\cite{second} & Lidar & 73.1 & 25.2 & 30.5 & 31.5 & 8.5 & 59.3 & 21.7 & 4.9 & 18.0 & 43.3 & 31.6  & 46.8\\
\hline
PP~\cite{pointpillar} & Lidar & {68.4} & 23.0 & 28.2 & 23.4 & 4.1 & 59.7 & 27.4 & 1.1 & 30.8 & 38.9 & 30.5  & 45.3 \\
\hline
Painting~\cite{Vora2019PointPaintingSF} & Lidar\&RGB &  \underline{77.9} &  \underline{35.8} &  \underline{36.1} &   \underline{37.3} & \textbf{15.8} & \textbf{73.3} &  \underline{41.5} &  \underline{24.1} & \bf{62.4} & \bf{60.2} & \bf{46.4} & \bf{58.1} \\
\hline
SSN & Lidar & \bf{80.7} & \bf{37.5} & \bf{39.9} & \textbf{43.9} &  \underline{14.6} &  \underline{72.3} & \bf{43.7} & {20.1} &  \underline{54.2} &  \underline{56.3} & \underline{46.3} & \underline{56.9} \\
\bottomrule
\end{tabular*}
\end{center}
\label{tab:nuscenes}
\vspace{-1ex}
\end{table*}

\noindent{\textbf{Results on nuScenes dataset.}}~~~ In this experiment, we test our model on nuScenes dataset and report the performance on the test set from official evaluation server. The results are shown in Table.~\ref{tab:nuscenes}. We give the detailed AP of each category and other metrics. It can be found that SSN achieves about 15\% improvement in mAP and 10\% in NDS compared to these lidar-based methods, even for some small objects, such as pedestrian and traffic cone. Even compared with the Lidar\&RGB fusion method~\cite{Vora2019PointPaintingSF}, our lidar-based model also achieves comparable performance and performs better in the main categories of traffic scenarios, such as Car, Truck, Bus and Motorcycle, \etc~
Note that the results of PointPillar and Painting~\cite{Vora2019PointPaintingSF} are copied from the original papers and for Second, we re-implement it under our setting and hyper-parameters are followed with SSN. 
For bicycle, due to its sparsity and low height, it is difficult to specify in the point cloud while it can be accessed in the image, thus the result of Bicycle in image detection is better than the 3D detection.

\noindent{\textbf{Results on Lyft dataset.}}~~~ For Lyft dataset, there is no official split of training set and validation set. Hence, we report the results on Kaggle competition (30\% test data is used for public validation but the host does not provide the ground truth. We submit the outputs of SSN and our baseline model to obtain the results). As Lyft dataset is a very new dataset, there is no official implementation. We re-implement PointPillar and Second to perform experiment on Lyft dataset, and optimization method and anchor matching strategy follow the SSN.
Table.~\ref{tab:lyft} shows the results of SSN and other existing methods on the test set. SSN consistently achieves the better performance with about 5\% improvement compared to existing methods.  Due to the strict metric (mAP-3D under IoU 0.5 to 0.95), the result on Lyft dataset is lower than nuScenes. Note that we only report the results of single model with single-scale training. The result on the official websites is 18.1\% which is applied with multi-scale training.

\begin{figure*}[t]
\begin{center}
   \includegraphics[width=1.0\linewidth]{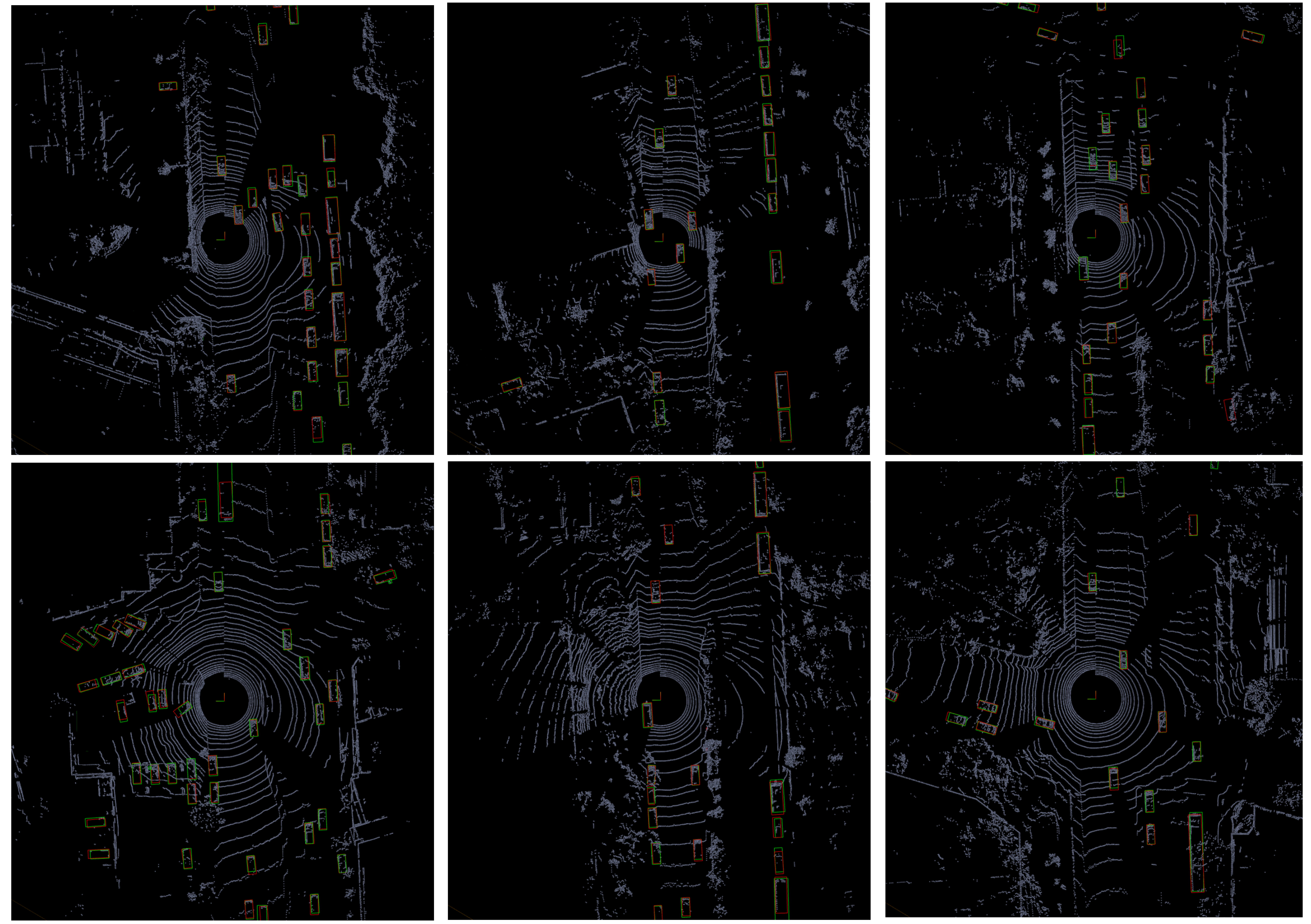}
\end{center}
\vspace{-4ex}
   \caption{We show the qualitative analysis of our model. These point clouds are sampled from Lyft dataset. Green boxes are the ground truth and red boxes are the model's predictions. To give a clear visualization, we crop the point could with range [-50, 50]
   (\textbf{Best viewed in color)}.}
\label{fig:lyft}
\end{figure*}
\begin{figure*}[t]
\begin{center}
   \includegraphics[width=0.8\linewidth]{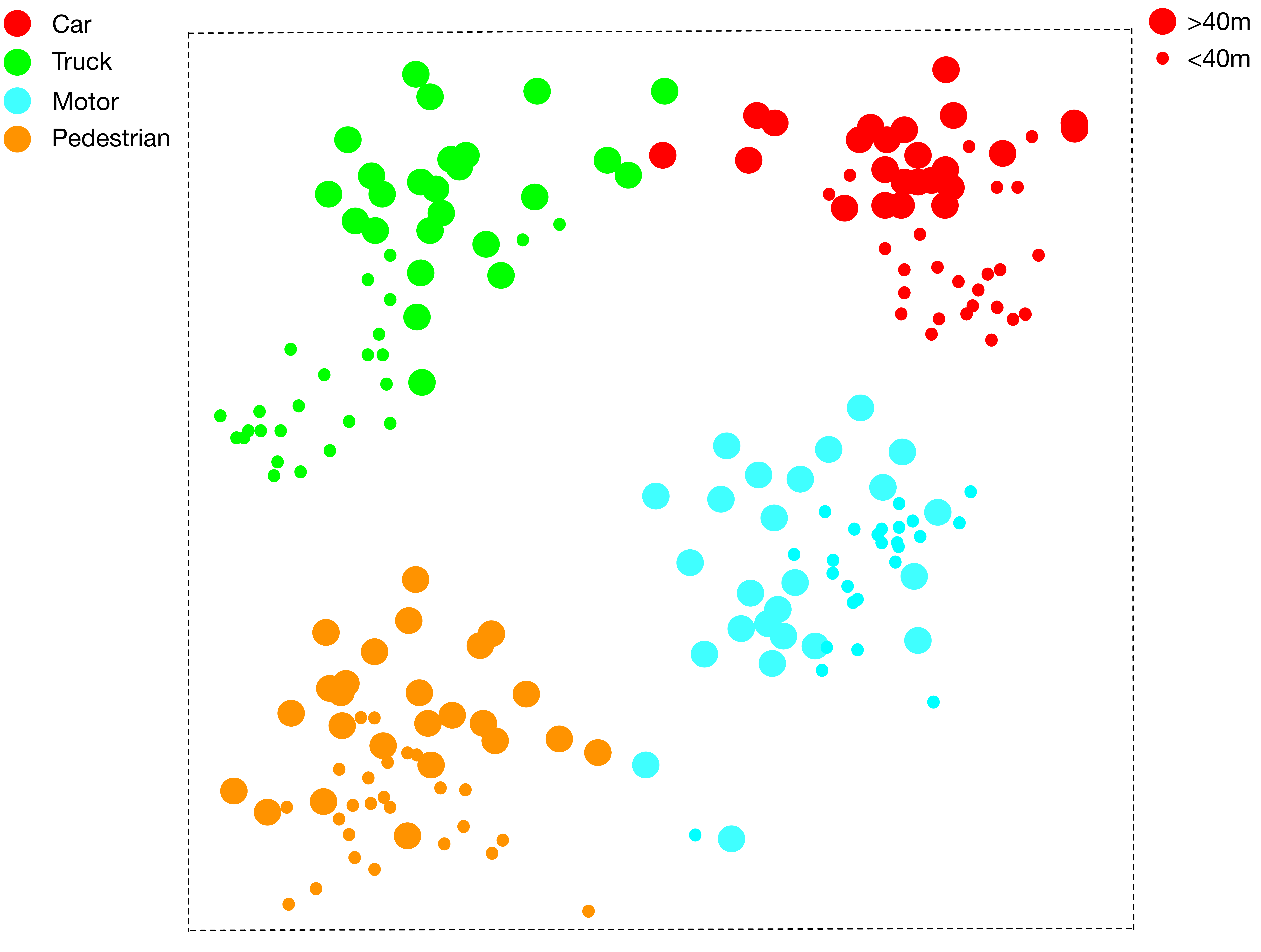}
\end{center}
\vspace{-4ex}
   \caption{We show the distribution of our shape signature via TSNE. 
   (\textbf{Best viewed in color)}.}
\label{fig:tsne}
\end{figure*}

\noindent{\textbf{Qualitative Analysis.}}~~~ We show several samples from the challenging Lyft dataset in Figure.~\ref{fig:lyft}. For ease of interpretation, we show the 3D boxes from the BEV perspective. It can be found that the car, bus and other vehicle achieve the decent performance. Some false positives and missing objects appear on the far range (about 50m). \textbf{TSNE visualization.} We use the TSNE to visualize the distribution of shape signature in Figure.~\ref{fig:tsne}. Four categories in nuScenes, including Car, Truck, Motorcycle and Ped, are sampled to display for a clearly visual effect. We sample 50 instances for each category, where 25 of them are with distance $<$ 40 meters and others are with distance $>$ 40 meters. It can be observed that the discrepancy across different classes is clear, which indicates the capability of our shape signature to separate the shape distribution across different categories.
Meanwhile, the distribution of shape signature within the same class differs with different distance (points with distance $<$ 40m and points with distance $>$ 40m cluster at different regions accordingly), which demonstrates the shape signature acts as a soft (not hard) constraint and keeps the shape distribution consistent (not same).

\subsection{Ablation Studies}
In this section, we perform the thorough ablation experiments to investigate the effect of different components in our method, including shape-aware grouping heads and shape signature, the scalability of the proposed shape signature with various backbone networks, and comparison with other shape signature.

\begin{minipage}[t!]{\textwidth}
        \begin{minipage}[t]{0.4\textwidth}
            \makeatletter\def\@captype{table}\makeatother
            \caption{Results on test set of Lyft dataset}
            \small
            \begin{tabular*}{1.0\linewidth}{c|c|c}
            \toprule
            Methods & Modality & mAP-3D \\
            \hline
            \hline
            Voxelnet~\cite{voxelnet} & Lidar & 10.1 \\
            \hline
            PointPillar~\cite{pointpillar} & Lidar & \underline{13.4}\\
            \hline
            Second~\cite{second} & Lidar & 13.0 \\
            \hline
            SSN & Lidar &\bf{17.9}\\
            \bottomrule
            \end{tabular*}
            
            \label{tab:lyft}
        \end{minipage}
        \hspace{.1in}
        \begin{minipage}[t]{0.5\textwidth}
            \makeatletter\def\@captype{table}\makeatother
            \caption{Experimental results of ablation studies on two key components on nuScenes dataset}
            \small
            \begin{tabular*}{1.0\linewidth}{c|c|c}
            \toprule
            Methods  & mAP & NDS \\
            \hline
            \hline
            PointPillar~\cite{pointpillar} & 29.4 & 44.9 \\
            \hline
            $+$ Shape-aware Grouping Heads & 40.6 & 51.3\\
            \hline
            $+$ Shape Signature&  45.3 & 57.0 \\
            \bottomrule
            \end{tabular*}
            
            \label{tab:ab1}  
        \end{minipage}
    \end{minipage}

\noindent{\textbf{Effect of Different Components.}}~~~ In this experiment, we choose the PointPillar as the backbone, and perform the ablation study by adding the components step-by-step. Due to the limited submissions in the evaluation server, we report the results on the official validation set of nuScenes dataset. As shown in Table.~\ref{tab:ab1}, it can be found that two key components, shape-aware grouping heads and shape signature, achieve the significant performance gain, with 6.4\% and 5.7\% improvements in NDS respectively, which demonstrates that the shape information does improve the multi-class detection.

\noindent{\textbf{Scalability of Shape Signature.}}~~~ To investigate the scalability of the proposed shape signature, we perform a thorough study, where the shape signature is combined with different backbone networks and tested on different datasets.  The detailed results are shown in Table.~\ref{tab:ab2}. For different backbone networks, we use PointPillar and Second, which utilize the 2D convolution and 3D convolution networks, respectively, and cover the mainstream in 3D object detection. It can be found that the shape signature could greatly improve the performance for different backbone networks on various datasets. 
Furthermore, it also achieves the consistent performance gain across different datasets, \ie~ nuScenes, Lyft and Kitti~\cite{kitti} dataset. Note that the mAP-3D in Lyft is similar to COCO dataset, which is much difficult than mAP in nuScenes and Kitti.
From these two perspectives, we can find that the proposed shape signature does possess good scalability and the exploration of shape information does improve the capability of detection networks in the discrimination of multiple categories.

\begin{table}[ht]
    \centering
    \caption{Experimental results of ablation studies on the scalability of shape signature. We perform the ablation with different backbones (PointPollar and Second) on three datasets (nuScenes, Lyft and Kitti). Note that for nuScenes, we report the results on the validation set and for Lyft, we report the results on the public test set in Kaggle. For Kitti, we report the moderate mAP with IoU=0.7 on two categories (car and pedestrian). ``PP" denotes PointPillar and ``SS" denotes our shape signature}
    \vspace{-2ex}
    \label{tab:ab2}
    \begin{tabular}{c|l|c|c|c|c|c|c|c}\toprule
    Dataset&Methods  & mAP & NDS & Dataset & mAP-3D & Dataset & mAP@car & mAP@ped\\
    \midrule
    \multirow{4}{*}{nuScenes}& PP\cite{pointpillar} & 29.4 & 44.9 & \multirow{4}{*}{Lyft} & 13.4 & \multirow{4}{*}{Kitti} & 74.3 & 41.9\\
    \cline{2-4}\cline{6-6}\cline{8-9}
    &\textbf{$+$ SS} & \textbf{36.6} & \textbf{49.8} & & \textbf{16.2} & & \textbf{76.2} & \textbf{43.5}\\
    \cline{2-4}\cline{6-6}\cline{8-9}
    &Second\cite{second} &  31.1 & 46.9 & & 13.0 & & 73.7 & 42.6\\
    \cline{2-4}\cline{6-6}\cline{8-9}
    &\textbf{$+$ SS} &  \textbf{34.3} & \textbf{48.9} & & \textbf{15.4} & & \textbf{75.4} & \textbf{44.1}\\
    \bottomrule
    
    \end{tabular}
\end{table}

\begin{table}
\small
\caption{Experimental results of Shape-aware grouping heads \emph{v.s.} One-to-one heads and Implicit shape signature \emph{v.s.} our shape signature. O-to-O Heads and SG Heads denote the one-to-one heads and shape-aware grouping heads, respectively}
\vspace{-2ex}
\setlength{\tabcolsep}{3.0pt}
\centering
\vspace{1ex}
\begin{tabular*}{1.0\linewidth}{l|c||c|c||c|c}
\toprule
Methods  & PP~\cite{pointpillar} & PP $+$ O-to-O Heads & \textbf{PP $+$ SG Heads} & PP $+$ IS\cite{IS} & \textbf{PP $+$ SS}\\
\hline
\hline
 mAP & 29.4 & 32.0 & \textbf{39.1} & 31.4 & \textbf{36.6} \\
\hline
 NDS & 44.9 & 46.2 & \textbf{51.0} & 46.7 & \textbf{49.8}\\
\bottomrule
\end{tabular*}
\label{tab:ab3}
\end{table}

\noindent{\textbf{Shape-aware Grouping Heads {\em{v.s.}} One-to-One Heads.}}~~~ To verify the effectiveness of the shape-aware grouping heads, we compare the shape-aware heads to the one-to-one heads, in which each head covers one category. The difference between two types of heads is the shape information investigation. From the results shown in Table.~\ref{tab:ab3}, it can be found that the shape-aware grouping heads perform much better than one-to-one heads in both metric terms, which further demonstrates the shape information benefits the multi-class discrimination. Moreover, the shape grouping strategy is also more effective than the one-to-one strategy, which groups the objects with similar shape and scale to aid the exploration of shape information.

\noindent \textbf{Comparison with other Shape Signature.}~~~ The previous work~\cite{IS} provides an implicit shape representation for instance segmentation. We adapt this approach into the point cloud segmentation and obtain the implicit shape signature with same dimension (``IS" is the notation). We compare the ``IS" with our shape signature (``SS") in Table.~\ref{tab:ab3}. It can be found that our shape signature outperforms the implicit shape signature with a large margin because ``SS" better handles difficulties from point cloud by completion and robustness enhancement. 
\noindent \textbf{Dimension of Shape Signature.}~~~ We use top 3 coefficients of Chebyshev approximation, because they principally and effectively cover the shape function. For example, for the bird-view shape vector of a car (we show full coefficients), [1.93, -0.65, 0.083, 4.68e-03, 1.064e-05, $\dots$], it can be found that top 3 coefficients contain the main knowledge and are appropriate as objective.

\section{Conclusion}
In this paper, we design a novel shape signature which acts as a soft constraint, and thus aid the feature capability of multi-class discrimination. Two appealing properties are carried, \ie~compact and effective as the objective and robust against the sparsity and noise. Based on the proposed shape signature, we develop the shape signature networks for object detection from point clouds, which makes use of shape information to promote the multi-class detection, through shape-aware heads and shape signature objective. We conduct extensive experiments and ablation studies, which demonstrate our model achieves
state-of-the-art and the proposed shape signature keeps
good scalability on various backbones.

%
%
\bibliographystyle{splncs04}
\bibliography{egbib}
\end{document}